%% file: main.tex
\documentclass{article} 
\usepackage{iclr2025_conference,times}

\input{math_commands.tex}

\usepackage{hyperref}
\usepackage{url}

\usepackage{xspace}
\usepackage{dsfont}
\usepackage{afterpage}

\usepackage{enumitem}
\usepackage{booktabs}
\usepackage{caption} 
\usepackage{multirow} 
\usepackage{enumitem}
\usepackage{colortbl} 
\usepackage{bbm}
\usepackage{algorithm}
\usepackage{algpseudocode}                  
\usepackage{setspace}                       
\usepackage{lipsum} 
\usepackage{subcaption} 

\usepackage{mathtools} 
\usepackage{bm} 
\usepackage{soul} 
\usepackage{nicefrac}
\usepackage{csquotes}

\usepackage{duckuments}
\usepackage{wrapfig}

\definecolor{cornellred}{rgb}{0.7, 0.11, 0.11}
\definecolor{cadmiumgreen}{rgb}{0.0, 0.42, 0.24}
\definecolor{aliceblue}{rgb}{0.91, 0.94, 0.97}
\definecolor{darkblue}{rgb}{0.83, 0.89, 0.97}
\definecolor{Red7}{rgb}{0.941, 0.243, 0.243}
\definecolor{Green7}{RGB}{55, 178, 77}
\definecolor{Blue9}{rgb}{0.098,0.3,0.9}

\usepackage{pifont}

\sethlcolor{aliceblue}

\hypersetup{
  linkcolor = cornellred,
  citecolor  = cadmiumgreen,
  colorlinks = true,
  urlcolor = Blue9
}

\newif\ifunderreview
\underreviewfalse 

\iclrfinalcopy

\title{Training-Free Representation Guidance for Diffusion Models with a Representation Alignment Projector}

\author{
\hspace{-0.25em}\textbf{Wenqiang Zu}\textsuperscript{1,3,4}\;
\textbf{Shenghao Xie}\textsuperscript{2}\;
\textbf{Bo Lei}\textsuperscript{4}\;
\textbf{Lei Ma}\textsuperscript{2}\thanks{Corresponding author: \texttt{lei.ma@pku.edu.cn}}\\
\textsuperscript{1}Institute of Automation, Chinese Academy of Sciences\;
\textsuperscript{2}Peking University\\
\textsuperscript{3}University of Chinese Academy of Sciences\;
\textsuperscript{4}Beijing Academy of Artificial Intelligence
}

\newcommand*{\ShowNotes}{} 

\ifdefined\ShowNotes
  \newcommand{\colornote}[3]{{\color{#1}\bf{#2: #3}\normalfont}}
\else
  \newcommand{\colornote}[3]{}
\fi

\begin{document}
\maketitle

\begin{abstract}
Recent progress in generative modeling has enabled high-quality visual synthesis with diffusion-based frameworks, supporting controllable sampling and large-scale training. Inference-time guidance methods such as classifier-free and representative guidance enhance semantic alignment by modifying sampling dynamics; however, they do not fully exploit unsupervised feature representations. Although such visual representations contain rich semantic structure, their integration during generation is constrained by the absence of ground-truth reference images at inference. This work reveals semantic drift in the early denoising stages of diffusion transformers, where stochasticity results in inconsistent alignment even under identical conditioning. To mitigate this issue, we introduce a guidance scheme using a representation alignment projector that injects representations predicted by a projector into intermediate sampling steps, providing an effective semantic anchor without modifying the model architecture. Experiments on SiTs and REPAs show notable improvements in class-conditional ImageNet synthesis, achieving substantially lower FID scores; for example, REPA-XL/2 improves from 5.9 to 3.3, and the proposed method outperforms representative guidance when applied to SiT models. The approach further yields complementary gains when combined with classifier-free guidance, demonstrating enhanced semantic coherence and visual fidelity. These results establish representation-informed diffusion sampling as a practical strategy for reinforcing semantic preservation and image consistency.
\end{abstract}

\input{sections/1_intro}

\input{sections/2_prelim}

\input{sections/3_method}

\input{sections/4_exps}
\input{sections/5_related}

\input{sections/conclusion}

\bibliography{iclr2025_conference}
\bibliographystyle{iclr2025_conference}

\appendix
\input{sections/appendix}

\end{document}

%% file: math_commands.tex
\usepackage{amsmath,amsfonts,bm}

\def\eqref#1{equation~\ref{#1}}

\def\1{\bm{1}}

\DeclareMathAlphabet{\mathsfit}{\encodingdefault}{\sfdefault}{m}{sl}
\SetMathAlphabet{\mathsfit}{bold}{\encodingdefault}{\sfdefault}{bx}{n}



%% file: sections/1_intro.tex
\begin{figure*}[h]
\vspace{-.5em}
\begin{minipage}{0.50\linewidth}
\includegraphics[width=0.99\textwidth]{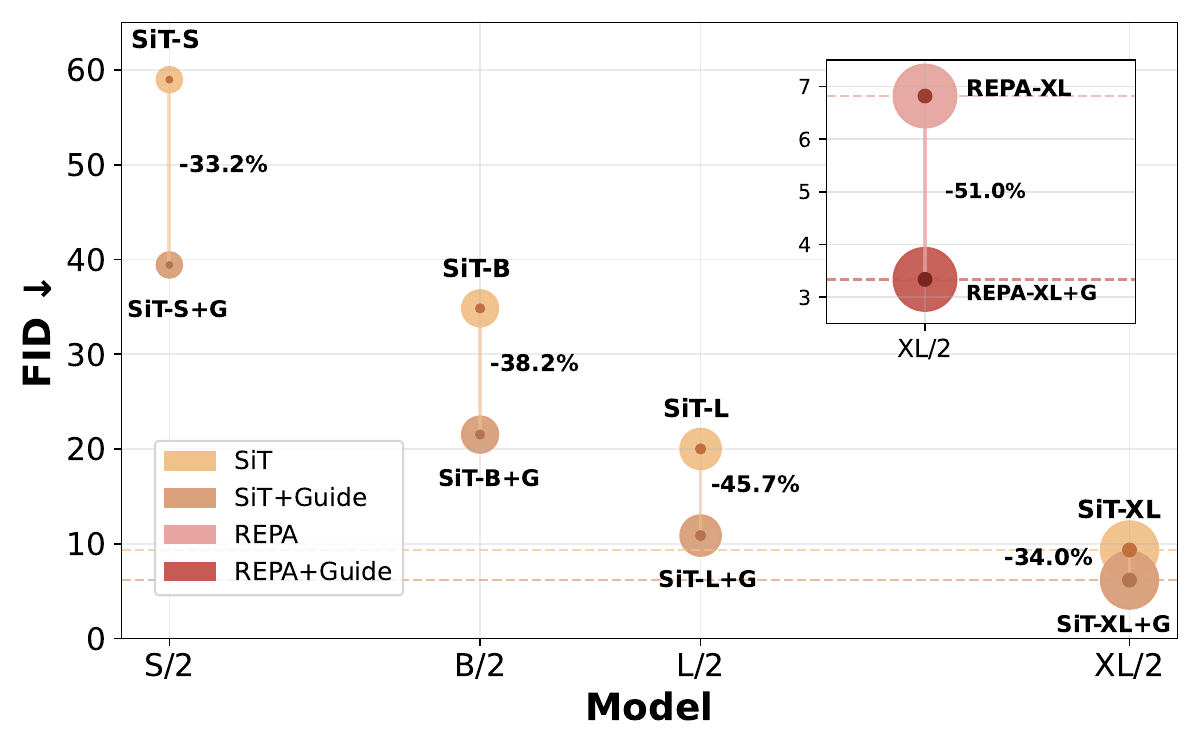}
\end{minipage}
\begin{minipage}{0.49\linewidth}
\includegraphics[width=0.19\linewidth]{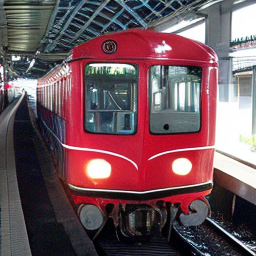}
\includegraphics[width=0.19\linewidth]{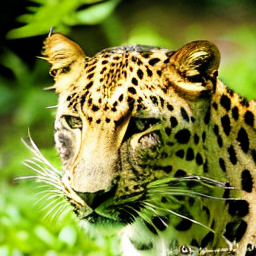}
\includegraphics[width=0.19\linewidth]{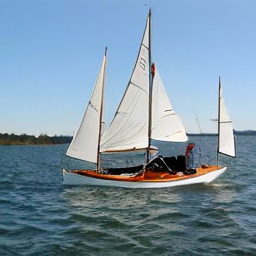}
\includegraphics[width=0.19\linewidth]{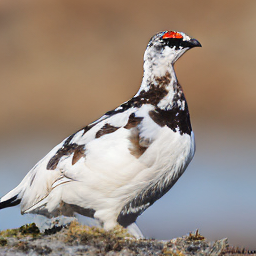}
\includegraphics[width=0.19\linewidth]{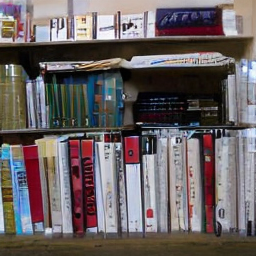}

\caption{Class-conditional generation on ImageNet 256×256 without CFG using ODE sampling. “+G” denotes that the corresponding SiT/REPA model uses guidance. Applying Representation Guidance yields significantly better generation quality than state-of-the-art diffusion/flow models.}
\label{fig1}
\end{minipage}
\end{figure*}

\section{Introduction}
\label{sec:intro}

Generative models have recently achieved significant progress in modeling complex visual data~\cite{ho2020denoising}. By generating images through progressive refinement or structured invertible flows~\cite{albergo2022building,lipman2022flow}, these approaches offer stable training, scalable architectures, and flexible control over the synthesis process. Advances in model design further enhance their expressive power~\cite{peebles2023scalable,ma2024sit}, enabling high-fidelity generation across a variety of settings. As a result, these models provide a versatile and effective framework for a wide range of image generation and conditional tasks.

A complementary line of research aims to enhance pretrained diffusion models during inference through various guidance techniques~\cite{bansal2023universal,ho2022classifier,kynkaanniemi2024applying,dinh2023pixelasparam,dinh2023rethinking,dinhrepresentative}. These methods introduce additional objectives into the sampling process, guiding generation toward the desired semantic or structural properties. Universal guidance~\cite{bansal2023universal}, for instance, incorporates the gradient of a chosen loss at each sampling step, directing the model’s updates toward the manifold defined by the guidance signal. Classifier-free guidance~\cite{ho2022classifier} adopts a different approach: by jointly training conditional and unconditional components, it interpolates between them at inference to improve semantic consistency while maintaining sample diversity. Representative guidance~\cite{dinhrepresentative} further biases sampling toward features characteristic of a target class, promoting fidelity to prototypical structures and patterns.


In addition to standard guidance strategies, there is growing interest in harnessing learned representations to actively shape the generative process~\cite{li2024return,yu2024representation,zheng2025diffusion}. Studies on diffusion transformers indicate that high-level visual features obtained through self-supervised learning serve as strong inductive biases, accelerating training and improving the fidelity of generated samples~\cite{yu2024representation}. These findings suggest that representations capture rich semantic and structural information that could, in principle, guide generation. However, translating this insight to inference is far from straightforward. Unlike during training, no ground-truth target image is available at the inference stage, leaving the model without an explicit reference to follow. Incorporating learned representations as guidance requires mechanisms that leverage these features while maintaining the flexibility of generation, highlighting a key frontier in connecting internal representations with the generative process.

To address this limitation, we note that effective guidance depends on a target representation, which is inherently unavailable during inference. Our observations indicate that diffusion generation achieves better alignment with the intended image’s semantics when augmented with representations predicted by a representation alignment projector. Motivated by this, we introduce a lightweight guidance mechanism that injects representation information from the target into each denoising step. This allows diffusion transformers to more consistently preserve and reflect the target semantics, enhancing fidelity and reliability while maintaining the model’s flexibility.

We motivate our approach through a detailed empirical study of contemporary diffusion transformers, including SiTs and REPAs, in combination with the self-supervised representation model DINOv2~\cite{oquab2024dinov}. Our analysis shows that in the initial denoising steps, when the signal is heavily corrupted by noise, generated samples often diverge from the semantic content encoded in the training data (Figure~\ref{fig3}). Even under identical class conditioning, different random seeds produce outputs with varying alignment to the target semantics, indicating limited stable guidance from learned representations. Incorporating an explicit semantic reference can help maintain coherence with the intended content, and external self-supervised representations can serve as such a reference during generation.

Building on these insights, we evaluate our method across diffusion transformer families including SiTs and REPAs, observing consistent improvements in class-conditional ImageNet generation. Representation-guided sampling reduces FID scores across SiT models of various scales, while REPA-XL/2 improves from 5.9 to 3.3, as shown in Figure~\ref{fig1}. Compared to state-of-the-art guidance techniques, our approach outperforms Representative Guidance on SiT models (FID 2.08 vs. 2.11). When combined with classifier-free guidance, these gains are further enhanced, reflecting improved semantic consistency and visual quality.

\begin{itemize}
    \item Integrating intermediate representations during inference provides a semantic prior that guides generation toward target-aligned content.
    \item We propose a representation-based guidance mechanism, \textbf{R-pred}, to enhance diffusion inference with minimal architectural changes.
    \item The method consistently improves diffusion transformer performance in class-conditional ImageNet generation, with REPA achieving an FID reduction from 5.9 to 3.3.
\end{itemize}

\begin{figure*}[!tbp]
    \includegraphics[width=\textwidth]{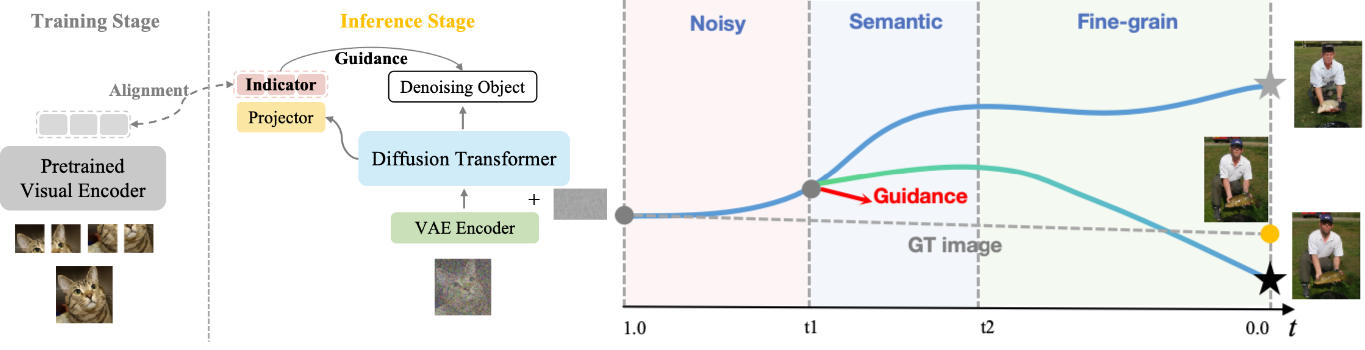} 
\caption{Overview of the Representation Guidance method. Left: A representation projector from a pretrained model (e.g., REPA) is used as an indicator to evaluate and guide the current denoising result via correlation. Right: Guidance is applied only within an intermediate timestep range to enhance semantic fidelity, without disrupting the initial coarse sampling or the final refinement stage.}
\label{fig2}
\end{figure*}

%% file: sections/2_prelim.tex
\section{Preliminaries}
\label{sec:preliminary}

Let $x_0 \sim p_{\mathrm{data}}$ denote a clean data sample, and let $x_1 \sim \mathcal{N}(0,I)$ denote a noisy sample independently drawn from a standard Gaussian distribution. We consider a continuous transform from $x_0$ to $x_1$ defined by an affine conditional flow
\begin{equation}
x_t = \psi_t(x_0 \mid x_1) := a_t x_0 + b_t x_1,\qquad t\in[0,1],
\end{equation}
where $a_t,b_t$ are smooth scalar schedules satisfying $a_0=1,\ b_0=0$ and $a_1=0,\ b_1=1$. This induces a path of intermediate marginal distributions $\{p_t\}_{t\in[0,1]}$ interpolating continuously from the clean data distribution $p_{\mathrm{data}}$ at $t=0$ to the Gaussian noise distribution at $t=1$.

The evolution of $x_t$ is governed by a deterministic flow described by the ODE
\begin{equation}
\mathrm{d}x_t = v_t(x_t)\,\mathrm{d}t,
\end{equation}
where $v_t:\mathbb{R}^d\!\to\!\mathbb{R}^d$ is a time-dependent velocity field. By the definition of the flow map $\psi_t$, the velocity field satisfies
\begin{equation}
v_t(x_t) = \dot{\psi}_t\!\left(\psi_t^{-1}(x_t)\right),
\end{equation}
ensuring that the marginal of $x_t$ equals $p_t$ for all $t\in[0,1]$. Once $v_t$ is learned, sampling is performed by solving the reverse-time ODE starting from $x_1\!\sim\!\mathcal{N}(0,I)$.

For the affine conditional flow, the conditional velocity field has the closed-form expression
\begin{equation}
v_t(x_t \mid x_1) = \dot{a}_t\,x_0 + \dot{b}_t\,x_1,
\end{equation}
where $x_t = a_t x_0 + b_t x_1$. Although the unconditional velocity field $v_t(x_t)$ involves integration over $x_0$, it can be approximated using conditional flow matching.

In practice, a neural network $v_\theta(x,t)$ is trained to approximate $v_t(x_t)$ by minimizing the conditional flow matching loss, 
where $x_0\sim p_{\mathrm{data}}$ and $x_1\sim\mathcal{N}(0,I)$:
\begin{equation}
\mathcal{L}_{\mathrm{CFM}}(\theta)
=
\mathbb{E}_{t\in[0,1]}
\!\left[\,\|\,v_\theta(x_t,t)-v_t(x_t\mid x_1)\|_2^2\,\right].
\end{equation}

A key result from conditional flow matching theory is that minimizing $\mathcal{L}_{\mathrm{CFM}}$ yields a consistent estimator of the unconditional velocity field $v_t(x_t)$, enabling direct ODE generation via
\begin{equation}
\mathrm{d}x_t = v_\theta(x_t,t)\,\mathrm{d}t.
\end{equation}

The above formulation accommodates arbitrary smooth schedules $\{a_t,b_t\}$. A commonly used special case is the linear (rectified) flow
\begin{equation}
a_t = 1-t,\qquad b_t = t,
\end{equation}
which results in a simple conditional velocity field $v_t(x_t\mid x_1)=x_1-x_0$. More general nonlinear schedules or discretized flows, including those used in diffusion models, can be recovered by suitable choices of $(a_t,b_t)$.

%% file: sections/3_method.tex
\begin{figure*}[htbp]
    \includegraphics[width=\textwidth]{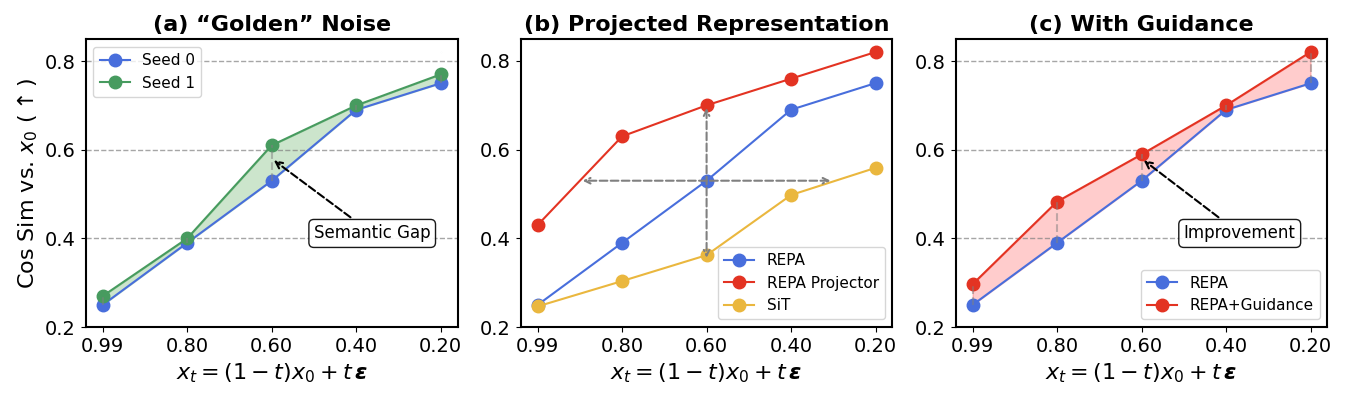} 
\caption{Observation of representation similarity of generated results (with a reference image). We randomly selected an image from class 0 of ImageNet and applied varying noise levels: $x_t = (1-t)x_0 + t \cdot \text{noise}$. (a) We observed the denoising process for different seeds and note that each target image has an associated noise pattern that best aligns with it, which can be considered as the "golden" noise for that target. (b) We examined the denoising outputs of pretrained generative models (SiT-XL/2, REPA-XL/2) and the representations directly predicted by REPA's projector, comparing their similarity to the reference image. (c) Using the projector-predicted representations to guide the generative model.}
\label{fig3}
\end{figure*}

\section{Method}

\subsection{Problem Setup}

We consider a pretrained continuous-time generative flow model defined by a velocity field
\begin{equation}
v_\theta : \mathbb{R}^d \times [0,1] \to \mathbb{R}^d,
\end{equation}
which governs the evolution of samples along a deterministic trajectory. Let $x_1 \sim \mathcal{N}(0,I)$ denote the initial noise state. The model generates a sample approximating the clean data $x_0 \sim p_{\mathrm{data}}$ by integrating the flow backward in time:
\begin{equation}
\tilde x_0 = \mathrm{ODESolve}\big(v_\theta, x_1, t\in[0,1]\big).
\end{equation}

To guide the generative process toward a desired target, a differentiable non-negative cost function $\mathcal{J}$ is defined to quantify the discrepancy between the generated output $\tilde x_0$ and a reference target $x_{\mathrm{ref}}$. The guidance problem can then be formulated as an optimization over the entire trajectory $\{x_t\}_{t\in[0,1]}$:
\begin{equation}
\min_{\{x_t\}_{t\in[0,1]}} \;\mathcal{J}(\tilde x_0, x_{\mathrm{ref}}),
\end{equation}
with the objective of adjusting the trajectory such that the final output aligns with the task-specific target.

Trajectory guidance is implemented via gradient-based updates to the intermediate states:
\begin{equation}
x_t \;\gets\; x_t - \alpha \, \nabla_{x_t} \mathcal{J}(\tilde x_0, x_{\mathrm{ref}}),
\end{equation}
where $\alpha>0$ controls the guidance strength. Since $\mathcal{J}$ depends on the terminal state $\tilde x_0$, gradients are backpropagated through the generative dynamics. This framework generalizes prior guidance approaches: some modify the trajectory directly, while others perform optimization in the denoised latent space and propagate updates to earlier states via deterministic flow integration.

If a reference sample $x_{\mathrm{ref}}$ is available, a natural choice for the cost function is the squared Euclidean distance:
\begin{equation}
\mathcal{J}(\tilde x_0, x_{\mathrm{ref}}) = \|\tilde x_0 - x_{\mathrm{ref}}\|_2^2,
\end{equation}
which enforces the generated output to closely match the reference.

In the absence of an explicit reference, the guidance objective can be formulated by estimating an implicit target $\hat x_0$ from the current state $x_t$ and using it to steer the trajectory. This can be expressed concisely as
\begin{equation}
\boxed{
\min_{\{x_t\}_{t\in[0,1]}} \; \mathcal{J}\big(\tilde x_0, \hat x_0(x_t)\big), \quad 
\hat x_0(x_t) = \mathcal{F}(x_t)
}
\end{equation}
where $\mathcal{F}$ denotes a function that infers an implicit target from the intermediate state $x_t$. In this representation-guided setting, each sample guides itself toward a plausible reconstruction, enabling trajectory-based generation without relying on an explicit reference.

\begin{figure*}[!tbp]
\centering
    \includegraphics[width=1.0\linewidth]{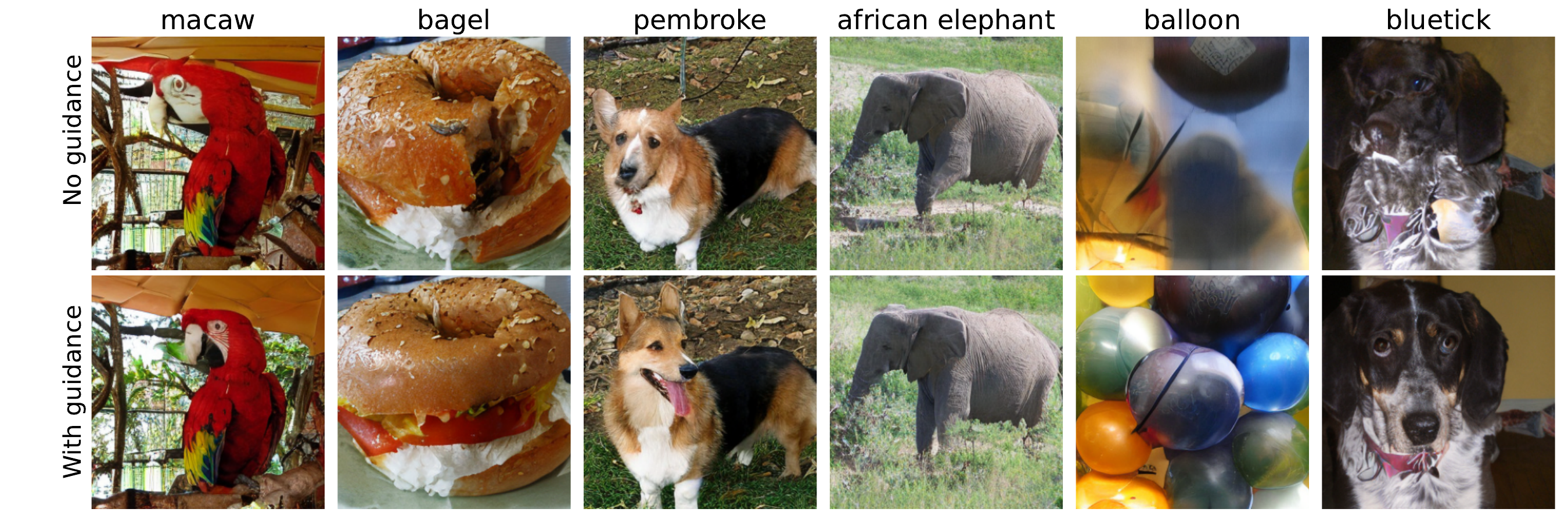} 
\caption{Selected samples on ImageNet 256$\times$256 from the REPA-XL/2 model (top) and with guidance (bottom). Applying Representation Guidance results in noticeable improvements in the generated images. We did not use classifier-free guidance, so $w = 1.0$.}
\label{fig4}
\end{figure*}

\subsection{Revisiting Implicit Targets in Flow Trajectories}

\subsubsection*{Motivation}
Continuous-time generative flows are trained under a well-defined supervision signal. Consider a clean data sample $x_0$ and an interpolated latent state
\begin{equation}
\boxed{
x_t = \underbrace{(1-t)x_0}_{\text{implicit }x_{\mathrm{ref}}} 
     + t x_1, 
\quad x_1 \sim \mathcal{N}(0,I)
}
\end{equation}

This interpolation implies that each intermediate state $x_t$ contains information about the clean target $x_0$, establishing a correspondence between the latent variables along the flow and their underlying reference. Consequently, the generative flow during inference is not target-free: each $x_t$ inherently points toward $x_0$, consistent with the velocity field learned during training.  

In practice, however, numerical integration errors and other factors can cause trajectories to drift, resulting in a reconstructed sample $\tilde x_0$ that deviates from the true target $x_0$. The objective is therefore to estimate a more accurate target for each intermediate latent state, ensuring that the generative flow remains coherently aligned with the original supervised objective.

\subsubsection*{Representation-Based Implicit Target}

As illustrated in Figure~\ref{fig2}, recent diffusion transformer variants~\cite{yu2024representation} commonly introduce a projector into the original architecture to align with representations from unsupervised pretraining, which is gradually becoming a default design choice. Let $\mathcal{F}$ denote a pretrained \emph{projector} that maps an intermediate latent $x_t$ to an estimate of the clean, unsupervised representation, e.g., a DINOv2 embedding. The predicted representation is
\begin{equation}
\hat{\phi}_t = \mathcal{F}(x_t),
\end{equation}
which approximates the representation of the underlying clean data from the current flow state $x_t$. Empirically, $\hat{\phi}_t$ provides a more reliable estimate of the clean signal than directly denoised latents, maintaining accuracy across noise levels and remaining resilient to trajectory drift during flow integration.

The implicit target for the flow is formalized via a representation-guided objective:
\begin{equation}
\boxed{
\mathcal{J}_{\mathrm{rep}}(x_t) = \|\phi(\hat{x}_0(x_t))  - \hat{\phi}_t\|_2^2, \quad
x_t \;\gets\; x_t - \alpha \, \nabla_{x_t} \mathcal{J}_{\mathrm{rep}}(x_t)
}
\end{equation}
where $\alpha>0$ controls the guidance strength. This update directs each intermediate latent $x_t$ toward regions of the latent space whose representations are consistent with the predicted clean embedding.

This approach ensures that the implicit target $\hat{\phi}_t$ is derived solely from the current latent $x_t$, without requiring access to the unknown clean sample $x_0$. Guidance is applied directly in latent space and integrates naturally with the continuous-time flow dynamics governed by the velocity field $v_\theta$. The method extends to arbitrary smooth interpolation schedules $\{a_t,b_t\}$, with representation-guided updates consistently applied at each intermediate time step.

\subsection{Multi-Step Representation Guidance}

R-Pred is formulated as a multi-step inference procedure that incorporates predicted clean embeddings into the generative trajectory, enhancing accuracy and consistency under continuous-time flow dynamics.

At each integration step $t_l$, the predicted representation 
\begin{equation}
\hat{\phi}_{t_l} = \mathcal{F}(x_{t_l})
\end{equation} 
serves as an estimate of the inaccessible clean target. The latent is updated to align with this predicted embedding:
\begin{equation}
x_{t_l} \;\gets\; x_{t_l} - \alpha \, \nabla_{x_{t_l}} 
\|\phi(\hat{x}_0(x_{t_l})) - \hat{\phi}_{t_l}\|_2^2
\end{equation}
where $\alpha>0$ controls the guidance strength.

For a discrete set of integration times $\{t_{low} < \dots < t_{high}\}$, the cumulative guidance objective can be expressed as
\begin{equation}
\mathcal{J}_{\mathrm{trajectory}} = 
\sum_{l=t_{\mathrm{high}}}^{t_{\mathrm{low}}} 
\|\phi(\hat{x}_0(x_{t_l})) - \hat{\phi}_{t_l}\|_2^2
\end{equation}
so that early steps encourage coarse alignment while later steps refine semantic and fine-grained details. The procedure is summarized in Algorithm~\ref{algo1}.

\begin{algorithm}[t]
\caption{Multi-Step Representation Guidance}
\begin{algorithmic}[1]
\Require Generative model $f_\theta$, representation encoder $\phi$, projector $\mathcal{F}$, integration times $\{t_0,\dots,t_T=1\}$, guidance interval $[t_\mathrm{low}, t_\mathrm{high}]$, initial latent $x_{t_T} \sim \mathcal{N}(0,I)$, guidance scale $\alpha$
\Ensure Final sample $\tilde x_0$

\State Initialize $x_{t_T}$

\For{$l = T$ down to $1$}
    \If{$t_l \in [t_\mathrm{low}, t_\mathrm{high}]$}
        \State Compute predicted representation $\hat{\phi}_{t_l} \gets \mathcal{F}(x_{t_l})$
        \State Evaluate current representation $\phi_{t_l} \gets \phi(x_{t_l})$
        \State Compute guidance loss $\mathcal{J} \gets \|\phi(\hat{x}_0(x_{t_l})) - \hat{\phi}_{t_l}\|_2^2$
        \State Update latent $x_{t_l} \gets x_{t_l} - \alpha \, \nabla_{x_{t_l}} \mathcal{J}$
    \EndIf
    \State Propagate latent with solver: $x_{t_{l-1}} \gets x_{t_l} + \mathrm{SolverStep}(f_\theta, x_{t_l}, t_l \rightarrow t_{l-1})$
\EndFor

\State \Return $\tilde x_0 \gets x_{t_0}$
\end{algorithmic}
\label{algo1}
\end{algorithm}

%% file: sections/4_exps.tex
\section{Experiments}

\subsection{Experimental Setup}

\paragraph{Dataset}  
We perform all experiments on the ImageNet dataset, using images at $256\times 256$ resolution. For each model, 50,000 samples are generated to enable comprehensive evaluation.  

\paragraph{Baselines and Guidance Methods}  
To evaluate the proposed R-Pred method, we conduct experiments on two diffusion/flow models: SITs and REPAs. All models are initialized with pretrained weights, and their parameters remain frozen during sampling and guidance. Additionally, results from other guidance methods applied to ADM~\cite{dhariwal2021diffusion} and DIT~\cite{Peebles2022DiT} models are included for reference. We consider state-of-the-art guidance strategies, including PxP~\cite{dinh2023pixelasparam}, ProG~\cite{dinh2023rethinking}, EDS~\cite{zheng2022entropy}, and RepG~\cite{dinhrepresentative}, denoted as the “+” variants. We also evaluate SIT+CFG, corresponding to SIT with classifier-free guidance. Details of these models and methods are provided in Appendix.

\paragraph{Implementation Details}  
All experiments use pretrained SIT or REPA models, and the sampling process follows the standard default settings of these models. The number of function evaluations (NFE) is set to 250 timesteps. Although model parameters remain frozen during sampling, the intermediate states $x_t$ are updated. Optimization is performed using AdamW, with an initial learning rate of $6\times 10^{-3}$ for SITs and $6\times 10^{-4}$ for REPAs. Guidance is applied once per sampling step. All experiments are conducted on four NVIDIA H100-80G GPUs. Additional experimental details are provided in Appendix. 

\paragraph{Apply Classifier-free Guidance}
For classifier-free guidance (CFG), the guidance strength is set to $w=1.5$ for SITs. For REPA, our reproduction of CFG did not yield consistent results, so we only report experiments for REPA without CFG.  The guidance time interval is critical: early guidance, when the model state is mostly noise, can produce inaccurate representation estimates, while late guidance has limited effect on the semantic trajectory. Therefore, guidance is applied in the normalized timestep range $t \in [0.8, 0.9]$ for SITs and $t \in [0.6, 0.9]$ for REPA without CFG, and $t \in [0.7-\Delta t, 0.7]$ for SITs with CFG, where $\Delta t = 1/\mathrm{NFE}$. Additional experimental details are provided in Appendix.

\paragraph{Evaluation Metrics}
To quantitatively assess generative performance, we adapt the Fréchet Inception Distance (FID) as the primary metric.

\begin{table*}[t!]
\centering
\small

\begin{minipage}{0.48\textwidth}
\centering
\caption{Class-conditioned image generation results on ImageNet 256$\times$256 with REPA-XL/2, without classifier-free guidance (CFG).}
\vspace{-0.05in}
\resizebox{\textwidth}{!}{%
\begin{tabular}{lccccc}
\toprule
Model & Type & Params & FID$\downarrow$ & sFID$\downarrow$ & IS$\uparrow$ \\
\midrule
REPA-XL/2 & ODE & 675M & 6.82 & 5.78 & 151.17 \\
\textbf{+ R-pred (ours)} & ODE & 675M & \textbf{3.34} & \textbf{4.77} & \textbf{196.23} \\
\midrule
REPA-XL/2 & SDE & 675M & 5.89 & 5.66 & 158.11 \\
\textbf{+ R-pred (ours)} & SDE & 675M & \textbf{3.50} & \textbf{5.01} & \textbf{192.90} \\
\bottomrule
\end{tabular}}
\label{tab:repa_wo_cfg}
\end{minipage}
\hfill
\begin{minipage}{0.48\textwidth}
\centering
\caption{Class-conditioned image generation results on ImageNet 256$\times$256 with vanilla SITs, with classifier-free guidance (CFG).}
\vspace{-0.05in}
\resizebox{\textwidth}{!}{%
\begin{tabular}{lccccc}
\toprule
Model & Type & Params & FID$\downarrow$ & sFID$\downarrow$ & IS$\uparrow$ \\
\midrule
SIT-XL/2+CFG & ODE & 675M & 2.15 & 4.60 & 258.09 \\
\textbf{+ R-pred (ours)} & ODE & 675M & \textbf{2.08} & \textbf{4.53} & \textbf{260.80} \\
\midrule
SIT-XL/2+CFG & SDE & 675M & 2.06 & \textbf{4.49} & \textbf{277.50} \\
\textbf{+ R-pred (ours)} & SDE & 675M & \textbf{2.02} & 4.53 & 269.34 \\
\bottomrule
\end{tabular}}
\label{tab:sit_cfg}
\end{minipage}

\vspace{0.25cm}

\begin{minipage}{0.48\textwidth}
\centering
\caption{Class-conditioned image generation results on ImageNet 256$\times$256 with vanilla SITs, without classifier-free guidance (CFG).}
\vspace{-0.05in}
\resizebox{1\textwidth}{!}{%
\begin{tabular}{lccccc}
\toprule
     Model & Type & Params & FID$\downarrow$ & sFID$\downarrow$ & IS$\uparrow$  \\
\midrule
SIT-S/2 & ODE & 33M  & 58.97 & 8.95 & 23.34 \\
{\textbf{+ R-pred (ours)}} & ODE & 33M   & \textbf{39.03}& \textbf{6.41}& \textbf{36.40} \\
\midrule
SIT-B/2 & ODE & 130M  & 34.84 & 6.59 & 41.53 \\
{\textbf{+ R-pred (ours)}} & ODE & 130M   & \textbf{21.52} & \textbf{5.47} & \textbf{62.99} \\
\midrule
SIT-L/2 & ODE & 458M  & 20.01& 5.31& 67.76\\
{\textbf{+ R-pred (ours)}} & ODE & 458M & \textbf{10.87}& \textbf{5.03}& \textbf{103.06}\\
\midrule
SIT-XL/2 & ODE & 675M  & 9.35& 6.38& 126.06\\
{\textbf{+ R-pred (ours)}} & ODE & 675M  & \textbf{6.17}& \textbf{4.78}& \textbf{150.83}\\
\midrule
SIT-S/2 & SDE & 33M  & 57.64 & 9.05& 24.78 \\
{\textbf{+ R-pred (ours)}} & SDE & 33M   & \textbf{39.42}& \textbf{6.32}& \textbf{35.90} \\
\midrule
SIT-B/2 & SDE & 130M  &33.02 &6.46 &43.71 \\
{\textbf{+ R-pred (ours)}} & SDE & 130M   & \textbf{21.12}& \textbf{5.46}& \textbf{64.34} \\
\midrule
SIT-L/2 & SDE & 458M  & 18.79& \textbf{5.29}& 72.02 \\
{\textbf{+ R-pred (ours)}} & SDE & 458M & \textbf{10.19}& {5.33}& \textbf{107.28}\\
\midrule
SIT-XL/2 & SDE & 675M  & 8.26& 6.32& 131.65\\
{\textbf{+ R-pred (ours)}} & SDE & 675M  & \textbf{4.16}& \textbf{4.85}& \textbf{183.30}\\
\bottomrule
\end{tabular}%
}
\label{tab:sit_wo_cfg}
\end{minipage}
\hfill
\begin{minipage}{0.48\textwidth}
\centering
\caption{Evaluation on ImageNet at $256\times256$ resolution comparing R-Pred with leading generative methods. Symbol $\dagger$ indicates scores reported in the original papers, respectively. All other results are obtained by reproducing the methods from their released implementations. For SiT models, we use ODE sampling.
}
\vspace{-0.05in}
\resizebox{\textwidth}{!}{%
\begin{tabular}{lcccc}
\toprule
Model & FID$\downarrow$ & sFID$\downarrow$ & Prec$\uparrow$ & Rec$\uparrow$ \\
\midrule
ADM$\dagger$ & 10.94 & 6.02 & 0.69 & \textbf{0.63} \\
ADM + RepG$\dagger$ & 7.83 & 5.79 & 0.72 & 0.61 \\
\midrule
ADM-G$\dagger$ & 4.58 & 5.23 & 0.81 & 0.52 \\
ADM-G + EDS$\dagger$ & 3.96 & 5.00 & 0.82 & 0.52 \\
ADM-G + PxP$\dagger$ & 4.00 & 5.19 & 0.81 & 0.53 \\
ADM-G + ProG$\dagger$ & 4.53 & 5.08 & \textbf{0.85} & 0.49 \\
ADM-G + ProG + EDS$\dagger$ & 3.84 & 5.00 & 0.83 & 0.51 \\
\midrule
DIT-CFG$\dagger$ & 2.27 & 4.80 & 0.82 & 0.58 \\
DIT-CFG + RepG$\dagger$ & 2.17 & 4.59 & 0.80 & 0.60 \\
\midrule
SIT-CFG + RepG & 2.10 & 4.61 & 0.80 & 0.60 \\
\rowcolor{gray!55}SIT-CFG + \textbf{R-pred (ours)} & \textbf{2.08} & \textbf{4.53} & 0.82 & 0.60 \\
\bottomrule
\end{tabular}}
\label{tab:vs_sota}
\end{minipage}

\end{table*}

\subsection{Quantitative Improvement}

R-pred provides consistent and notable quantitative improvement across different backbones, sampling strategies, and CFG settings on ImageNet 256$\times$256, proving its generality and scalability from small SIT models to large transformer-based diffusion systems.

\paragraph{REPA without CFG.}
As shown in Table~\ref{tab:repa_wo_cfg}, integrating R-pred into REPA-XL/2 leads to clear performance gains. 
Under ODE sampling, FID decreases from 6.82 to 3.34 (51.0\% reduction), sFID from 5.78 to 4.77 (17.5\% reduction), and IS increases from 151.17 to 196.23 (+29.8\%). 
Under SDE sampling, FID improves from 5.89 to 3.50 (40.0\% reduction) and sFID from 5.66 to 5.01 (11\% reduction), indicating that R-pred benefits the denoising trajectory across different solvers.

\paragraph{SITs without CFG.}
Table~\ref{tab:sit_wo_cfg} reports improvements across all SIT scales. 
With ODE solvers, R-pred consistently reduces FID and increases IS, with larger models showing greater gains:
SIT-S/2 achieves 33.8\% lower FID and 55.9\% higher IS; 
SIT-B/2 shows a 38.2\% FID reduction and 51.6\% IS increase; 
SIT-L/2 reaches a 45.7\% FID reduction and a 52.1\% IS increase; 
SIT-XL/2 also improves with a 34.0\% FID reduction and 19.6\% IS increase. 
Under SDE sampling, similar trends hold. For instance, SIT-L/2 improves from 18.79 to 10.19 (45.8\% reduction) with IS increasing from 72.02 to 107.28 (+49.0\%), and SIT-XL/2 further reaches an FID of 4.16. 
These results indicate that R-pred mitigates error accumulation during generation and enhances semantic consistency in transformer-based diffusion models.

\paragraph{SITs with CFG.}
Even with classifier-free guidance, R-pred continues to improve sample quality (Table~\ref{tab:sit_cfg}). 
For SIT-XL/2+CFG under ODE, FID decreases from 2.15 to 2.08 and IS increases from 258.09 to 260.80. 
Under SDE, FID improves from 2.06 to 2.02, suggesting that R-pred further refines convergence and enhances visual fidelity in CFG settings.

\paragraph{Comparison to state-of-the-art.}
Table~\ref{tab:vs_sota} compares R-pred with recent generative models. 
When combined with SIT-CFG, our method reaches FID 2.08 and sFID 4.53, outperforming the SIT-CFG+RepG guidance method.
This demonstrates that R-pred provides an effective plug-in for improving fidelity, structural consistency, and distribution alignment without modifying the underlying architecture.

\subsection{Qualitative Results}  
Figure~\ref{fig4} illustrates the effect of the Representation Guidance method on class-conditional generation. For original generations with noticeable quality issues, such as `bagel' and `bluetick', the guided results show clear improvements in fidelity and visual quality. For images where the semantics are already well captured, guidance preserves the overall structure and content, producing only subtle, minor refinements. This demonstrates that Representation Guidance can correct errors while maintaining consistency for already accurate samples.

\begin{wraptable}[14]{r}{0.5\textwidth}
\centering\small
\vspace{-0.2in}
\caption{\textcolor{black}{Ablation study on the guidance time interval. $t=1$ corresponds to a fully noisy input, $t=0$ corresponds to a clean image, and $w$ denotes the classifier-free guidance scale. $\Delta t = 1/\mathrm{NFE}$. ODE-based sampling is employed.}}
\vspace{-0.1in}
\resizebox{0.48\textwidth}{!}{
\begin{tabular}{lcccc}
\toprule
     Model & Samples & $w$ & Interval & FID$\downarrow$ \\
     \midrule
     \rowcolor{gray!55}
     SIT-XL/2+R-pred & 5000  & 1.0 & [0.80, 0.90] & \textbf{11.57} \\
     SIT-XL/2+R-pred & 5000  & 1.0 & [0.60, 0.90] & 15.10 \\
     SIT-XL/2+R-pred & 5000  & 1.0 & [0.60, 0.80] & 13.26 \\
     SIT-XL/2+R-pred & 5000  & 1.0 & [0.20, 0.60] & 17.49 \\
     \midrule
     SIT-XL/2+R-pred & 5000  & 1.5 & [0.60, 0.80] & 9.82 \\
     SIT-XL/2+R-pred & 5000  & 1.5 & [0.60, 0.70] & 9.25 \\
     SIT-XL/2+R-pred & 5000  & 1.5 & [0.80-$\Delta t$, 0.80] & 9.10 \\
     \rowcolor{gray!55}
     SIT-XL/2+R-pred & 5000  & 1.5 & [0.70-$\Delta t$, 0.70] & \textbf{9.09} \\
     \midrule
     SIT-XL/2+R-pred & 50000 & 1.5 & [0.70-$\Delta t$, 0.70] & \textbf{2.08} \\
     \bottomrule
\end{tabular}
}
\label{tab:ab_timeinterval}
\end{wraptable}

\subsection{Ablation Study}

The choice of guidance time interval has a significant impact on the generated results. Excessive guidance can disrupt the generative trajectory, so it is preferable to apply guidance during an earlier denoising stage to facilitate semantic formation. Table~\ref{tab:ab_timeinterval} presents an ablation on the effect of different guidance intervals. For generation without CFG, we apply guidance over a continuous interval, whereas when both CFG and representation guidance are applied, we find that guiding only near a single timestep with a limited number of updates yields better results.

%% file: sections/5_related.tex
\section{Related Work}

\textbf{Diffusion and Flow Models.} Diffusion-based~\cite{ho2020denoising,song2020score} and flow-based~\cite{albergo2022building,lipman2022flow,liu2022flow} generative frameworks have expanded the scope of high-fidelity image synthesis. Flow-based models, which learn continuous invertible transformations, complement diffusion approaches by providing exact mappings and competitive likelihood estimation, together forming a versatile generative landscape. A line of research explores the use of external representations to guide generation~\cite{li2024return,yu2024representation,zheng2025diffusion}. REPA~\cite{yu2024representation} aligns representations within the VAE latent space, while RAE~\cite{zheng2025diffusion} replaces the VAE encoder with a DINOv2-based encoder. These studies show that features from large-scale self-supervised encoders improve semantic consistency and visual fidelity while accelerating convergence. Prior work also indicates that the standard noise-prediction objective requires regression over the full noise space, rather than the low-dimensional manifold of clean images, increasing capacity demands~\cite{li2025back}. Directly predicting the clean image $x_0$ leverages this low-dimensional structure and enhances early-stage denoising. Building on these insights, the \textbf{R-pred} method integrates clean-image structure with external representations, providing a representation-guided refinement mechanism that improves coherence and semantic fidelity in diffusion transformer generation.

\textbf{Guidance Methods.} Guidance mechanisms are widely used to improve the quality of diffusion model samples. Early approaches incorporate gradients from pretrained classifiers or multimodal models such as CLIP into the sampling process, directing generation toward semantically or perceptually preferred results and improving metrics like FID~\cite{bansal2023universal,dinh2023pixelasparam,chefer2023attend,guo2024initno,grimal2025saga}. Recent methods include Reg~\cite{dinhrepresentative}, which is gradient-based, and RepG, which uses precomputed class centers for each ImageNet category as guidance targets. However, applying the same guidance to all samples of a class is overly coarse. In contrast, our approach predicts a sample-specific target for each noisy latent and guides generation toward it, with the predicted target remaining more accurate than the current noisy image throughout denoising.


%% file: sections/conclusion.tex
\section{Conclusion}
In this work, we show that integrating external self-supervised representations during inference provides a strong semantic prior for diffusion transformers, improving alignment and fidelity in class-conditional generation. The proposed \textbf{R-pred} guidance mechanism injects representation information into intermediate sampling steps, serving as an effective semantic anchor without altering model architecture or training. Extensive experiments across SiTs and REPAs validate its consistent performance gains, highlighting representation-guided sampling as a powerful and general strategy for enhancing diffusion inference.

Despite these advances, the approach remains gradient-based and requires computing representation loss during sampling, which introduces additional overhead. A promising future direction is to explore more efficient guidance paradigms that reduce gradient dependency, potentially by establishing explicit flows or mappings between latent variables and representation space. Constructing a unified representation-latent flow framework may further improve controllability, enable training-free semantic editing, and deepen understanding of how high-level features interact with generative trajectories.

%% file: sections/appendix.tex
\clearpage

\section{Hyperparameter and More Implementation Details}
\label{appen:main_setup}

\subsection{Further Implementation Details}

We implement our experiments using the REPA and SiT architectures as specified in Table~\ref{tab:hyperparam}. For all setups, input latent vectors are of dimension $32\times32\times4$, obtained via pre-trained stable diffusion VAE. For the REPA-Projector, we consistently use an 8-layer MLP with cosine similarity as the projection objective, and DINOv2-B as the representation encoder. Across all experiments, we optimize using AdamW with learning rates set per Table~\ref{tab:hyperparam}, without additional weight decay. We use mixed-precision training (fp16) and apply gradient clipping to stabilize optimization.

For the diffusion process, we employ linear interpolants $\alpha_t = 1-t$ and $\sigma_t = t$, with denoising performed via v-prediction. Sampling is done using either ODE or SDE solvers for 250 steps, and classifier-free guidance is applied with the CFG scale indicated in Table~\ref{tab:hyperparam}. Guidance intervals are set differently across tasks, ranging from fixed values to dynamic schedules as noted in Table~\ref{tab:hyperparam}. No explicit data augmentation is applied.

\begin{table*}[!t]
    \centering\small
    \caption{Hyperparameter setup.}
    \resizebox{\textwidth}{!}{%
    \begin{tabular}{l c c c c c}
        \toprule
         & {Figure 3} & Figure 4 & Table 1 & Table 2 & Table 3 \\
        \midrule
        \textbf{Architecture} \\
        Model size. & REPA-XL/2 & REPA-XL/2 & REPA-XL/2 & SiT-S,B,L,XL/2 & SiT-XL/2 \\
        Input dim. & 32$\times$32$\times$4 & 32$\times$32$\times$4 & 32$\times$32$\times$4 & 32$\times$32$\times$4 & 32$\times$32$\times$4 \\
        Num. layers & 28 & 28 & 28 & 12,12,24,28 & 28 \\
        Hidden dim. & 1,152 & 1,152 & 1,152 & 384,768,1,024,1,152 & 1,152 \\
        Num. heads & 16 & 16 & 16 & 6,12,16,16 & 16 \\ 
        \midrule
        \textbf{REPA-Projector} \\
        Projection depth & 8 &  8 & 8 & 8 & 8 \\
        $\mathrm{sim}(\cdot, \cdot)$ & cos. sim.  & cos. sim. & cos. sim. & cos. sim.  & cos. sim.  \\
        Representation Encoder $f()$ & DINOv2-B  & DINOv2-B & DINOv2-B & DINOv2-B  & DINOv2-B\\
        \midrule
        \textbf{Optimization} \\
        Guidance Interval & 0.99,0.80,0.60,0.40,0.20 & [0,60,0.90]& [0,60,0.90] & [0.60,0.80] & [0.7-$\Delta$t,0.7]  \\ 
        Optimizer & AdamW & AdamW & AdamW & AdamW & AdamW \\
        lr & 0.0006 & 0.0006 &  0.0006 & 0.0060 & 0.0060 \\
        \midrule
        \textbf{Interpolants} \\
        $\alpha_t$ & $1-t$ & $1-t$ & $1-t$ & $1-t$ & $1-t$ \\
        $\sigma_t$ & $t$ & $t$ & $t$ & $t$ & $t$ \\
        $w_t$ & $\sigma_t$ & $\sigma_t$ & $\sigma_t$ & $\sigma_t$ & $\sigma_t$ \\
        Denoising objective & v-prediction & v-prediction & v-prediction & v-prediction & v-prediction \\
        Sampler & ODE & ODE & ODE,SDE & ODE,SDE & ODE,SDE \\
        Sampling steps & 250 & 250 & 250 & 250 & 250 \\
        CFG scale & 1.0 & 1.0 & 1.0 & 1.0 &  1.5 \\
        \bottomrule
    \end{tabular}
    }
    \label{tab:hyperparam}
\end{table*}

\textbf{Sampler.}
We follow the sampling settings used in the original SiT paper~\cite{ma2024sit} and the REPA paper~\cite{yu2024representation} for both ODE and SDE sampling.
For SDE sampling, we adopt the Euler–Maruyama sampler with the diffusion coefficient set as $w_t = \sigma_t$, and the final SDE step size is fixed to 0.04.

\textbf{Reproduction Details of RepG}
We follow the settings in the RepG paper~\cite{dinhrepresentative} as closely as possible. For example, we set the number of representative vectors for each category to $K=5$, and we use cosine similarity loss. For the selection of representative vectors, we first compute the mean representation of each ImageNet class (i.e., the features encoded by Dinov2 for all images), then select $K$ ($K=5$) representations corresponding to the images whose features are closest to the class mean representation. Latents are obtained using the SD-VAE encoder. During guidance, we try two sampling strategies: randomly sampling one of the $K$ representative vectors, and always choosing the vector closest to the class mean. Based on the results, we adopt the strategy of randomly selecting a representative vector at every step. The guidance interval is kept consistent with the R-pred (ours) method. For $\text{cfg}=1.5$, the time interval is set to $[0.7-\Delta t,\,0.7]$, where $\Delta t = 1/\text{NFE}$.

\section{Running Time of Representation Guidance}
\begin{table}[h]
\centering
\caption{GPU time for sampling 50,000 images at a resolution of 256$\times$256 using 4 H100-80G GPUs.}
\begin{tabular}{l c c}
\hline
Model & Sampler & Computational Cost \\
\hline
No Guidance & ODE & 56m \\
Representation Guidance & ODE & 1h37m \\
CFG & ODE & 1h50m \\
CFG + Representation Guidance & ODE & 2h06m \\
\hline
CFG & SDE & 7h52m \\
CFG + Representation Guidance & SDE & 8h22m \\
\hline
\end{tabular}
\label{tab:time}
\end{table}

Table~\ref{tab:time} reports the overall sampling time, showing that Representation Guidance introduces moderate overhead compared to unguided generation and remains efficient under both ODE and SDE sampling.

\section{Pretrained Baseline Models}
\label{appen:baselines}
We evaluate our method using four representative generative backbones: ADM, DiT, SiT, and REPA.
\begin{itemize}[leftmargin=0.2in]
    \item \textbf{ADM}~\cite{dhariwal2021diffusion} extends diffusion modeling with an optimized U-Net architecture and introduces classifier-based guidance to enhance controllable sampling while maintaining sample diversity.
    \item \textbf{DiT}~\cite{Peebles2022DiT} replaces convolutional designs with a fully transformer-driven diffusion framework and integrates adaptive normalization (AdaIN-zero) modules for scalable high-resolution generation.
    \item \textbf{SiT}~\cite{ma2024sit} further advances transformer diffusion by reformulating training under continuous flow matching, enabling more efficient optimization and improved generative quality.
    \item \textbf{REPA}~\cite{yu2024representation} builds on DiTs/SiTs by incorporating representation alignment with a self-supervised encoder during training, employing an additional projector for feature matching, which leads to faster convergence compared with DiTs/SiTs.
\end{itemize}

\section{Comparison with Existing Guidance Methods}
We compare our approach with several recent inference-time guidance techniques: EDS, PxP, ProG, and RepG.
\begin{itemize}[leftmargin=0.2in]
    \item \textbf{EDS}~\cite{zheng2022entropy} introduces entropy-aware scaling for DDPMs, adaptively restoring conditional guidance during sampling and reducing overconfidence from classifier predictions.
    \item \textbf{PxP}~\cite{dinh2023pixelasparam} frames guidance in a gradient-based perspective, resolving conflicts between update directions using a projection strategy.
    \item \textbf{ProG}~\cite{dinh2023rethinking} presents Progressive Guidance, a generalization of classifier guidance that gradually incorporates class-level information during early sampling to improve both diversity and robustness.
    \item \textbf{RepG}~\cite{dinhrepresentative} is a gradient-based method that steers the generation trajectory toward class-representative vectors computed from ImageNet, using them as reference targets during sampling.
\end{itemize}

\section{Supplementary Related Work}
Additional research has explored guidance strategies beyond gradient-based methods. Classifier-free guidance (CFG)~\cite{ho2022classifier} steers generation away from the unconstrained unconditional distribution, emphasizing higher-probability conditional samples to improve output quality. Extensions such as AutoGuidance~\cite{karras2024guiding} enhance this approach by introducing a weaker auxiliary model whose predictions regularize the main model during sampling, mitigating low-quality outputs. Although effective, CFG-style methods require extra unconditional training, adding complexity. Another approach focuses on aligning guidance with the target sample’s semantics. Methods~\cite{zhou2025golden,bai2024zigzag,wang2025towards} generate semantically consistent “golden noise” via denoising or inversion and use these noise targets to guide the diffusion process toward more faithful samples.

\begin{figure*}[!tbp]
\centering
    \includegraphics[width=1.0\linewidth]{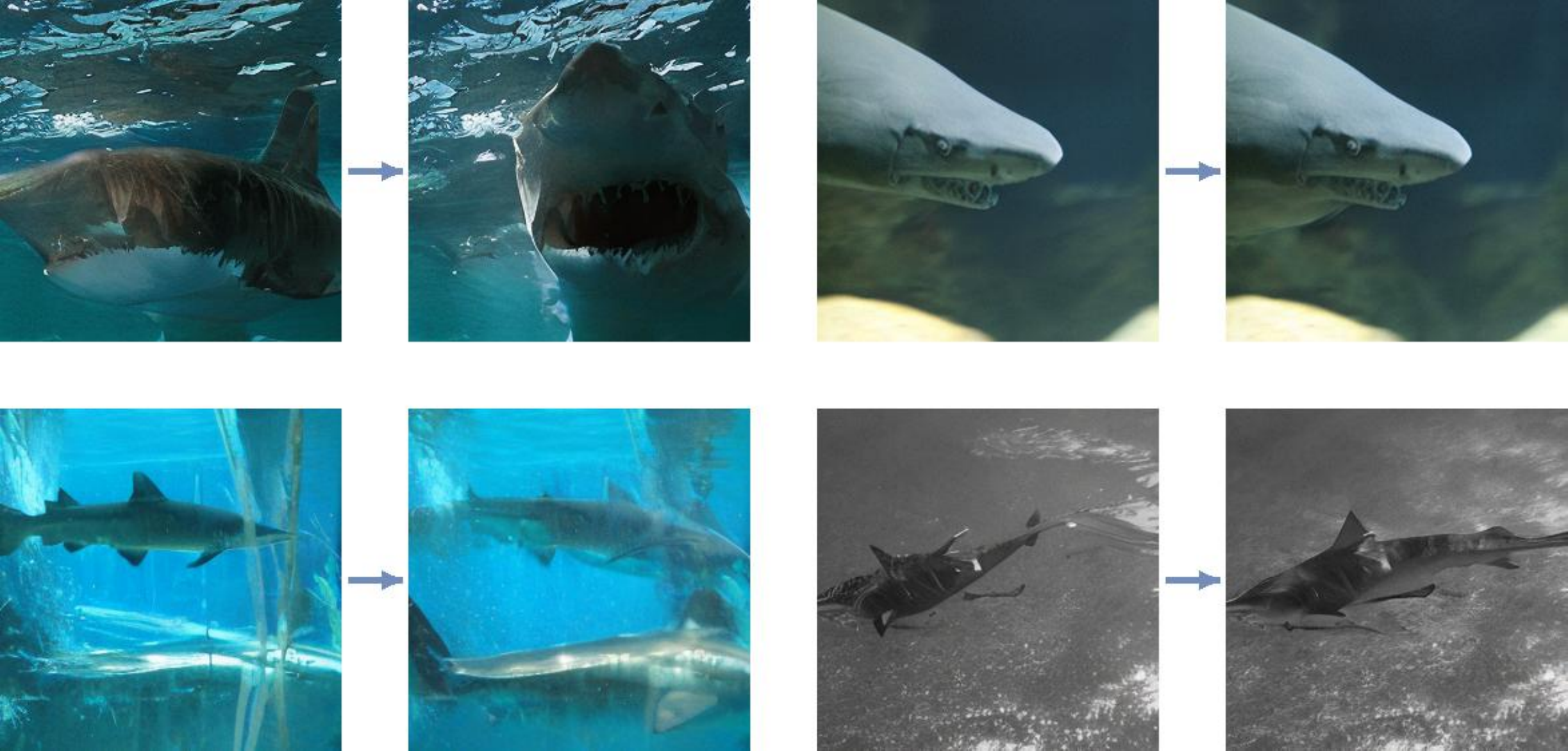} 
\caption{ImageNet256x256/class: tiger shark. The images on the left, preceding the arrow, represent incorrect outputs produced by REPA-XL/2 in the absence of CFG and other guidance, whereas the images on the right, following the arrow, illustrate the improvements obtained through R-pred guidance..}
\label{supp_3}
\end{figure*}

\begin{figure*}[!tbp]
\centering
    \includegraphics[width=1.0\linewidth]{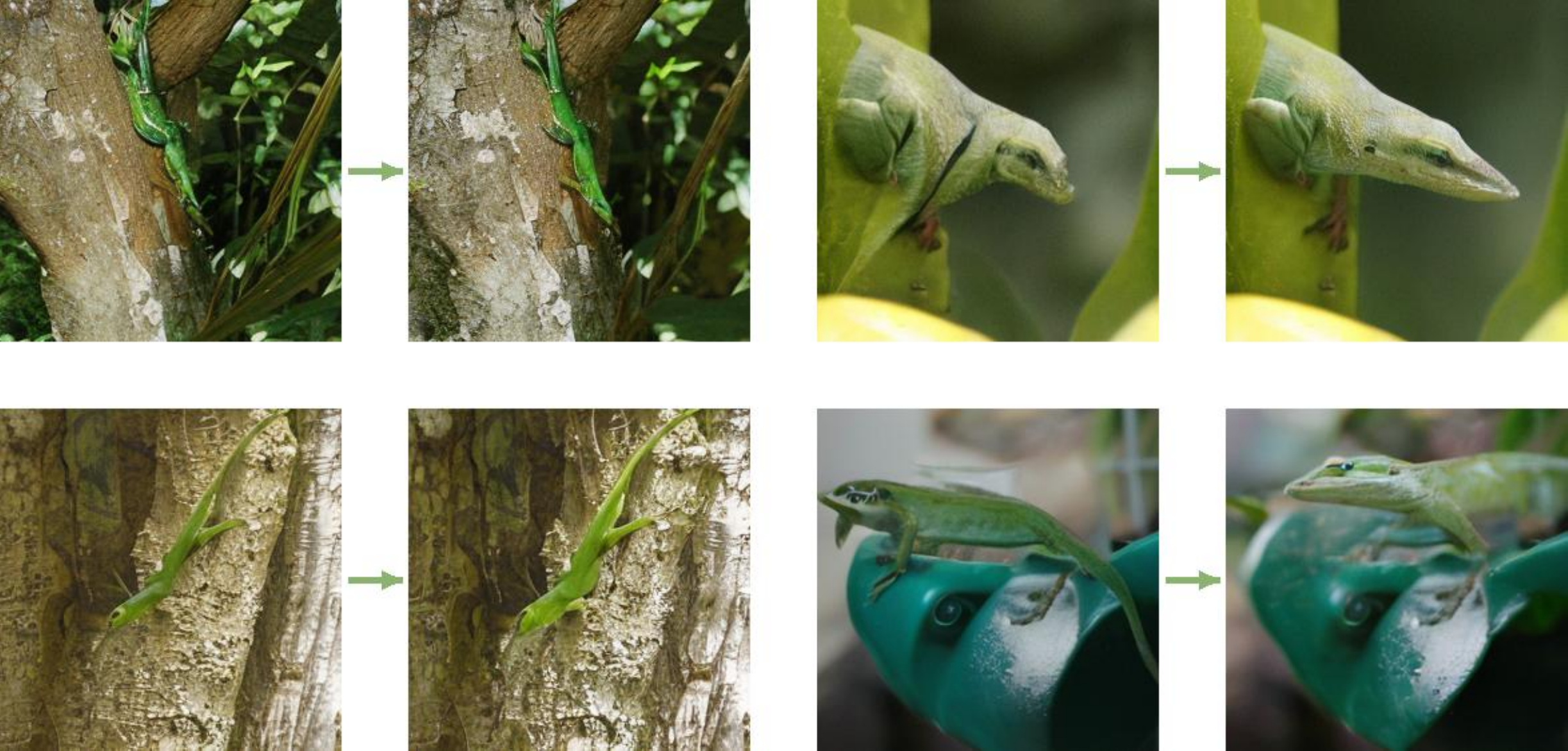} 
\caption{ImageNet256x256/class: green lizard. The images on the left, preceding the arrow, represent incorrect outputs produced by REPA-XL/2 in the absence of CFG and other guidance, whereas the images on the right, following the arrow, illustrate the improvements obtained through R-pred guidance..}
\label{supp_46}
\end{figure*}

\begin{figure*}[!tbp]
\centering
    \includegraphics[width=1.0\linewidth]{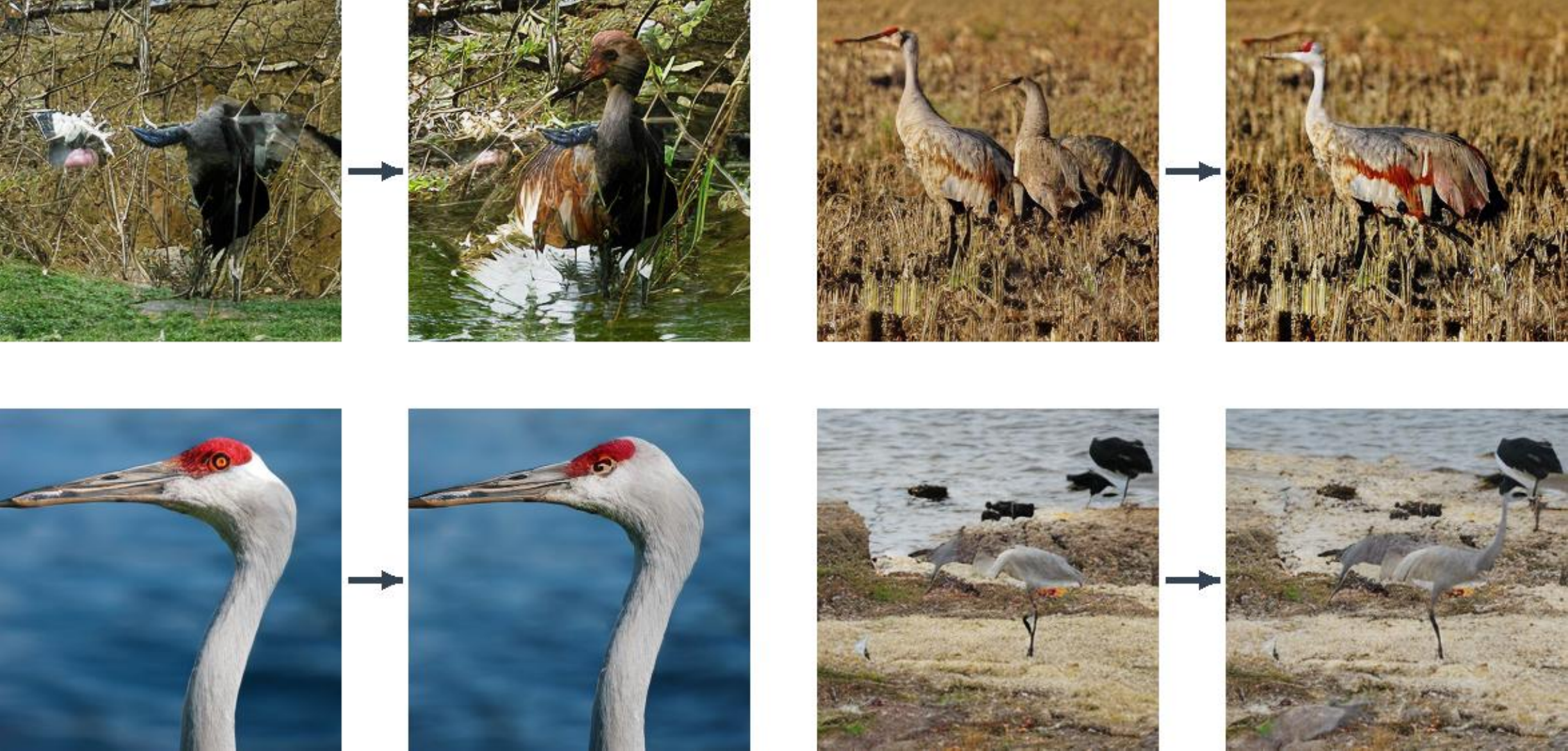} 
\caption{ImageNet256x256/class: crane. The images on the left, preceding the arrow, represent incorrect outputs produced by REPA-XL/2 in the absence of CFG and other guidance, whereas the images on the right, following the arrow, illustrate the improvements obtained through R-pred guidance..}
\label{supp_134}
\end{figure*}

\begin{figure*}[!tbp]
\centering
    \includegraphics[width=1.0\linewidth]{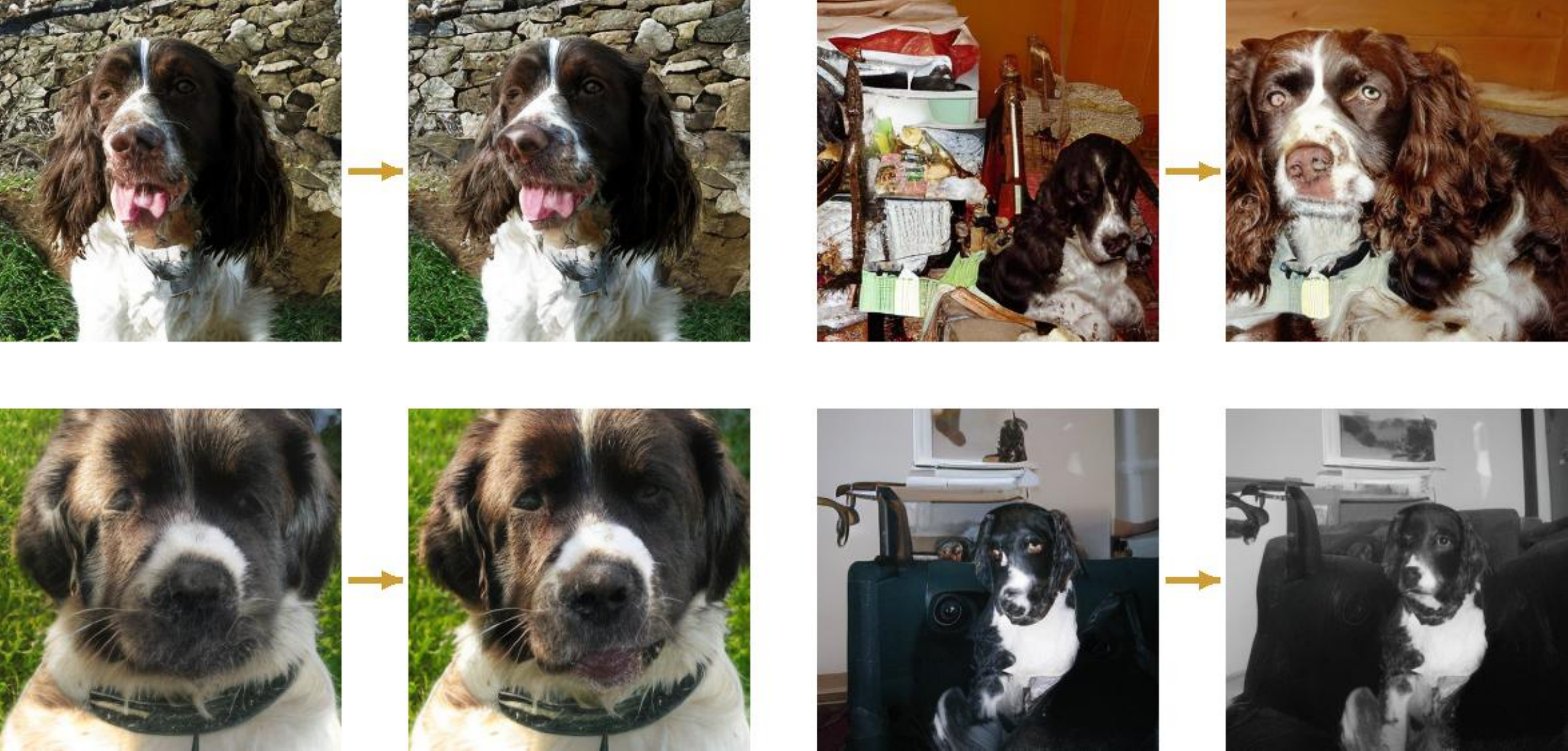} 
\caption{ImageNet256x256/class: English springer. The images on the left, preceding the arrow, represent incorrect outputs produced by REPA-XL/2 in the absence of CFG and other guidance, whereas the images on the right, following the arrow, illustrate the improvements obtained through R-pred guidance..}
\label{supp_217}
\end{figure*}

\begin{figure*}[!tbp]
\centering
    \includegraphics[width=1.0\linewidth]{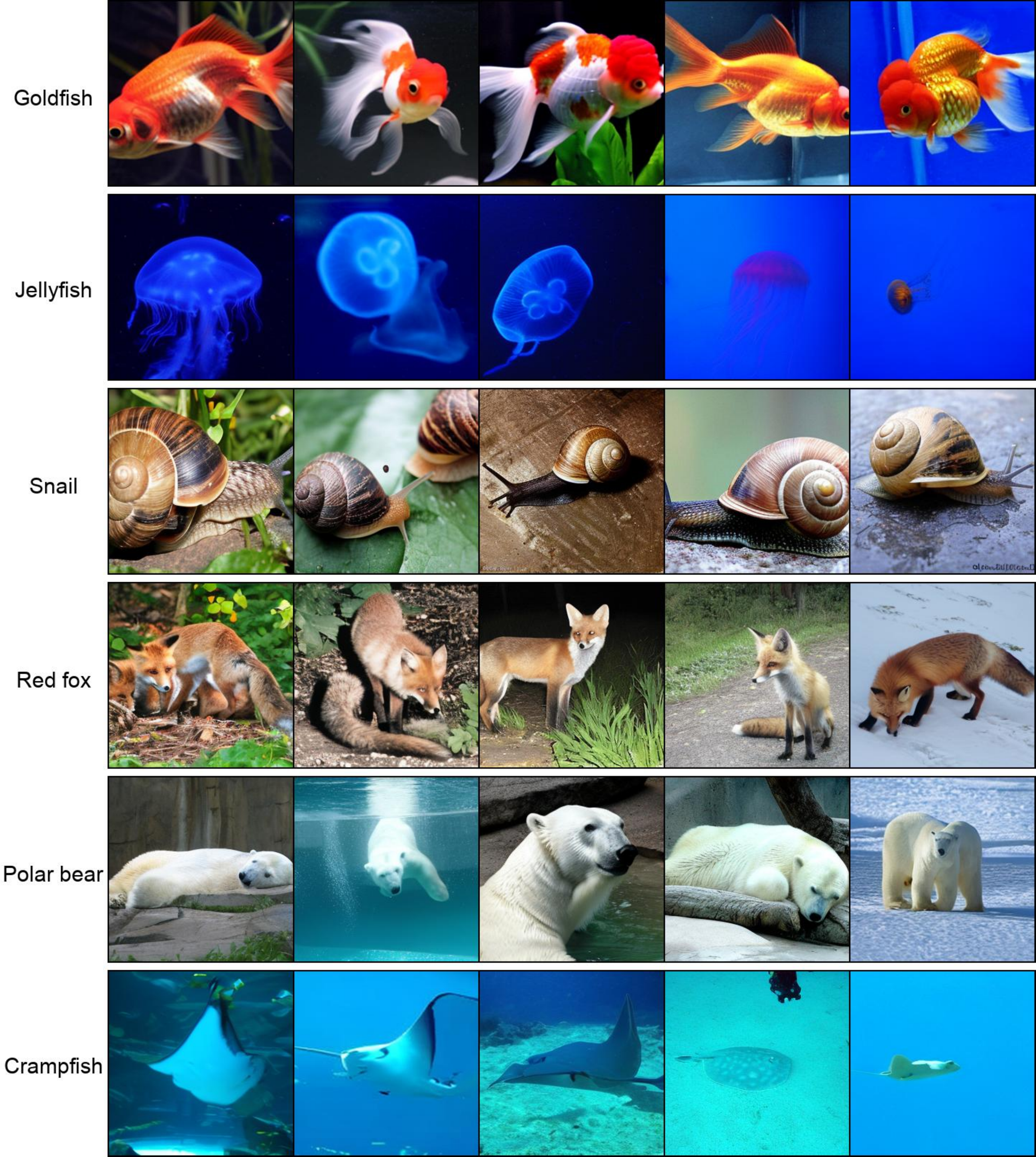} 
\caption{Samples generated by SiT-XL/2 using R-pred with a CFG scale of 4.0.}
\label{supp_all}
\end{figure*}